\documentclass[]{interact}

\usepackage{epstopdf}
\usepackage[caption=false]{subfig}

\usepackage[numbers,sort&compress]{natbib}
\bibpunct[, ]{[}{]}{,}{n}{,}{,}
\renewcommand\bibfont{\fontsize{10}{12}\selectfont}
\makeatletter
\def\NAT@def@citea{\def\@citea{\NAT@separator}}
\makeatother

\theoremstyle{plain}

\theoremstyle{definition}

\theoremstyle{remark}

\begin{document}

\articletype{ARTICLE TEMPLATE}

\title{Enhancing Student Engagement in Online Learning through Facial
Expression Analysis and Complex Emotion Recognition using Deep
Learning}

\author{
\name{Rekha R Nair\textsuperscript{a}\thanks{CONTACT R.~R.~N. Author. Email: rekhasanju.sanju@gmail.com}, Tina Babu\textsuperscript{a} and Pavithra K\textsuperscript{a}}
\affil{\textsuperscript{a}Department of Computer Science and Engineering, Alliance University, Bengaluru, India; 
}
}

\maketitle

\begin{abstract}
In response to the COVID-19 pandemic, traditional physical classrooms have transitioned to online environments, necessitating effective strategies to ensure sustained student engagement. A significant challenge in online teaching is the absence of real-time feedback from teachers on students learning progress. This paper introduces a novel approach employing deep learning techniques based on facial expressions to assess students engagement levels during online learning sessions. Human emotions cannot be adequately conveyed by a student using only the basic emotions, including anger, disgust, fear, joy, sadness, surprise, and neutrality. To address this challenge, proposed a generation of four complex emotions such as confusion, satisfaction, disappointment, and frustration by combining the basic emotions. These complex emotions are often experienced simultaneously by students during the learning session. To depict these emotions dynamically,utilized a continuous stream of image frames instead of discrete images. The proposed work utilized a Convolutional Neural Network (CNN) model to categorize the fundamental emotional states of learners accurately. The proposed CNN model demonstrates strong performance, achieving a 95\% accuracy in precise categorization of learner emotions.
\end{abstract}

\begin{keywords}
Facial Emotion Detection; CNN; Pattern
detection
\end{keywords}

\section{Introduction}
Covid 19 Pandemic has made all the educational schools across the world adapt teaching online\cite{bokhare2023emotion}. Distance learning remains a vital and ongoing process that provides essential support to both students and educators in their teaching and learning endeavors across the globe\cite{rashmi2023facial}. Technology and online learning materials can help students develop successful self-directed learning techniques\cite{tripathi2021visualization}. There are various obstacles in education, such as invigilation and learning coordination, as a result of the widespread adoption of distance learning in the modem world\cite{sathyamoorthy2023facial}. In higher education, distance learning has provided a useful substitute for conventional instruction. It might be challenging for university lecturers to comprehend the emotions and unusual behaviors of their pupils during class\cite{bakariya2023facial}. While online learning has achieved considerable success and popularity, it still faces a challenge in adapting pedagogical approaches in real-time based on the learner's evolving behavior and emotions, a capability that is more readily achievable in traditional face-to-face learning settings\cite{karilingappa2023human}. As a result, the learning process can become somewhat mechanized, which significantly influences the depth of knowledge acquisition. Conventional methods often rely on the analysis of facial expressions in photographs to gauge a learner's emotional state. However, it's crucial to recognize that human emotions are inherently intricate and multifaceted, extending beyond fundamental feelings such as anger, disgust, fear, joy, sadness, surprise, and neutrality\cite{rokhsaritalemi2023exploring}. However, it is possible to take into account a mixture of two or more emotions that may appear on the face over time\cite{HAARIKA2023338}. The four complex emotions that are a composite of fundamental human emotions such as confusion, satisfaction, disappointment, and frustration that a learner frequently experiences in concert throughout a learning session. Instead of using discrete pictures, the usage of a fixed set of continuous image frames to accurately represent these mixed feelings. To categorize the fundamental emotions and subsequently determine the learners' state of mind, called a CNN model. Convolutional neural networks (CNN) have helped a number of effective artificial intelligence algorithms, particularly deep learning algorithms, become well-known in the computer vision sector. It has often been used in image classification and recognition\cite{mohan2023face}\cite{nair2023multiresolution}. It is important to note that achieving a high level of accuracy in image processing is essential for the successful implementation of face detection and recognition systems. This precision is a fundamental requirement to ensure that the system is not only effective but also reliable in its performance.

This paper endeavors to introduce an enhanced face recognition approach with the primary objective of improving the effectiveness of emotion recognition. This advanced technique is designed to surpass the accuracy levels achieved by traditional methods. It leverages a combination of software techniques, computer vision algorithms, and deep learning models, specifically CNNs, to establish an innovative system. This system empowers educators with the capability to efficiently orchestrate classroom activities and enhance communication with their students during lessons, all while ensuring students' engagement and monitoring their behavioral state in the classroom. The main emphasis of this research is as follows:
\begin{itemize}
    \item To identify the basic facial emotions in learning session.
\item	To give the accurate combinational emotion detection for identified emotions.
\item	To detect learners state of mind accurately.
\end{itemize}

There are various obstacles in education, such as invigilation and learning coordination, because of the widespread adoption of distance learning in the modem world. Understanding students' emotions and unusual behavior during class sessions is a challenge for university instructors.  Detection of state of mind by facial expressions of online learners is difficult with models trained with basic emotions. To solve this problem, a new novel deep learning model which detects state of mind by combinatorial facial emotions using CNN algorithm to be developed.
\begin{figure}[!tbp]
  \centering
  \begin{minipage}[b]{0.55\textwidth}
    \includegraphics[width=\textwidth, height=7cm]{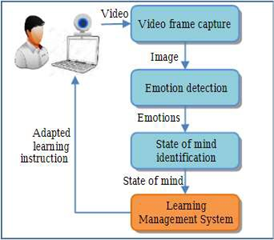}
    \caption{Proposed model for emotion recognition system.}
    \label{fig:Proposed model for emotion recognition system}
  \end{minipage}
  \hfill
  \begin{minipage}[b]{0.4\textwidth}
    \includegraphics[width=\textwidth, height=7cm]{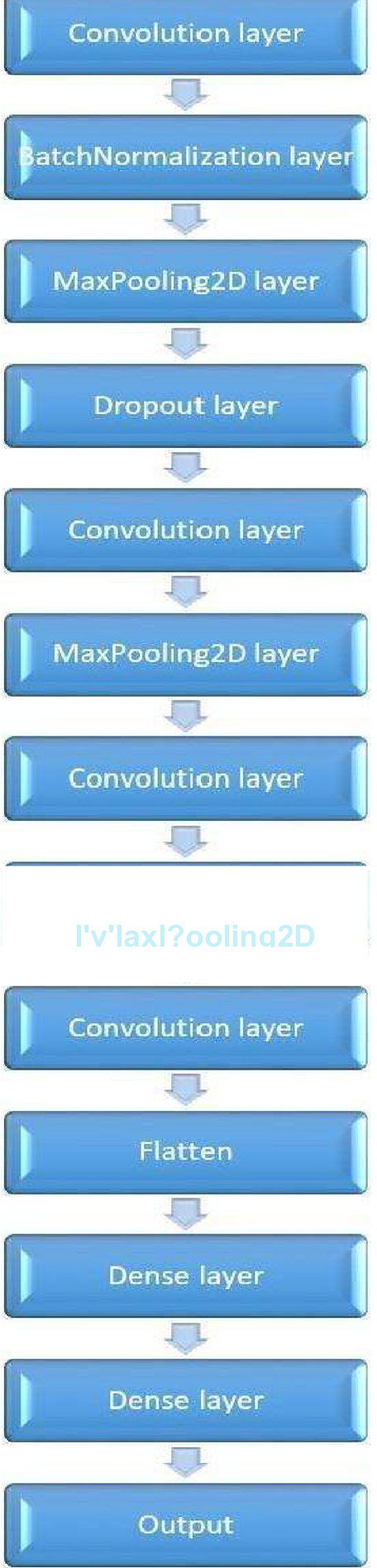}
    \caption{Proposed CNN model and its layered structure.}
  \label{fig:Proposed CNN model and its layered structure}
  \end{minipage}
\end{figure}
\section{Materials and methods}
\label{Materials and methods}
\subsection{Proposed Methodology}
Employing CNN as a deep learning system, this study utilizes them to process input images, assess and categorize different features and objects present in the images, and differentiate between them effectively. The CNN is harnessed to scrutinize real-time video frames with the purpose of predicting the probability of each of the seven core emotional states manifesting. Furthermore, the analysis model relies on real-time input derived from the CNN models' output data to discern the emotional states of students, offering a means of detecting their state of mind during educational interactions
\begin{itemize}
    \item \textbf{}Face detection is a pivotal initial step in the recognition process, achievable through classical cascade classifiers or deep learning methods. OpenCV often assists in this process. Grayscale conversion streamlines object detection, as color information is unnecessary. Deep learning classifiers yield detections with face coordinates and dimensions. In real-time facial recognition, enhancing accuracy is imperative. Online and distance learning aim to capture student emotions for recognition, offering potential for development and improvement in this domain
    \item \textbf{Facial Emotion Detection}
By matching facial expression patterns, emotion from the face picture might be identified. When it comes to recognizing facial expressions and subsequently categorizing them for emotion, machine learning technologies are intricate. In CNN-based techniques each feature map is employed within interconnected neural networks to identify facial expressions and assign them to corresponding emotion classes. It's worth noting that CNN demonstrates a higher level of accuracy in comparison to other neural network-based classifiers, making it a favorable choice for this purpose.
\item \textbf{Identification of Learners' State of Mind}
When a student is engaged in a learning process, it has been found that the learner's emotions alter gradually rather than abruptly. Additionally, during the learning process, the learner's face displays variations in emotion for a while (at least for 3 to 5 seconds). To gauge a learner's mental state while they are learning, a series of photos must be collected throughout time. For purposes of generalization, were assume that a change in human feeling may occur in around 6 seconds. To assign scores to each class of fundamental emotion, the facial expressions within each image are accurately recognized. The predominant emotion in the facial image receives a notably high confidence score. Different score values are computed for each specific emotion, as a single face may display a range of emotional nuances. This approach allows for a nuanced assessment of the emotions expressed.

In order to discern the array of emotions within an image, the emotion recognition module harnesses a pre-trained CNN classifier. By employing a sequence of six consecutive images as a "window frame," the state of mind detection module is capable of assessing the learner's emotional patterns over the preceding six seconds. The resulting emotional pattern provides insight into the learner's current mental state and emotional condition.

\end{itemize}
\subsection{Implementation}
The architectural layout of our CNN model is depicted in Figure \ref{fig:Proposed CNN model and its layered structure}. This structure comprises five convolution layers, each equipped with a Rectified Linear Unit (ReLu) activation function. Additionally, there are three pooling layers, two fully connected layers, and an output layer. The specific functionalities and parameter settings for each layer are detailed as follows:

\begin{itemize}
    \item In the convolution layer, the feature map is generated from the input images and is achieved by employing convolution kernels configured with a size of 3 x 3.
\item	The purpose of the max pooling layer is to reduce dimension of the data while retaining crucial features and patterns. Each max pooling layer is configured with a stride value of 2 and a pooling window size of 2x2. This design choice ensures effective dimensionality reduction while preserving significant information.
\item	The flatten layer transforms the 2-dimensional data into a 1-dimensional format, making it suitable for input to a fully connected layer. Conversion process allows for seamless integration into subsequent network components.
\item	The dense layer or the fully connected layer, acts as a collection of two or more interconnected neural networks. The 1-dimensional data obtained from the flattening layer is provided as input to the input nodes of each dense layer. This architecture allows for intricate neural network interactions, enhancing the model capacity to capture complex patterns and relationships within the data.
\item	The output layer is composed of seven nodes, each utilizing the SoftMax activation function. Each of these nodes corresponds to a distinct category of emotions, allowing the network to predict and classify different sets of emotions.
\end{itemize}

The CNN model is being implemented through the Python programming language. This constructed model is subsequently subjected to further training and testing using a facial expression dataset to assess its accuracy and performance
\subsection{Learners verification}
Emotions and states of mind are inherently subjective and often resist precise quantification or formal expression. Consequently, to accurately gauge the implicit state of mind in learners, it is imperative to obtain validation from the individuals involved. To assess the accuracy of our emotion model and the approach for recognizing emotional patterns, it is essential to validate the classified emotion patterns with the learners themselves. In this regard, the approach is undergoing assessment and validation involving 40 graduate-level course participants. This evaluation entails a brief online tutorial followed by a machine learning test session. During the learning session, video recordings of each candidate are taken at different time points to ascertain and understand the learner's state of mind.

Once the learning session concludes, the recorded video is meticulously analyzed frame by frame, aiming to extract the emotional patterns and, consequently, the learner's evolving state of mind over time. The mechanism for detecting emotion patterns operates at 6-second intervals to derive the learner's state of mind. These identified states of mind are then aggregated throughout the entire learning session. To validate the accuracy of this process, the candidates are invited to provide feedback regarding the correctness of the aggregated learner's state of mind as assessed. This hypothesis posits that the learners possess the capacity to accurately recognize and respond to their own states of mind.
\begin{table}
  \begin{minipage}{.6\columnwidth}
    \centering
    \caption{CNN's performance metric for detecting emotions}
\label{performance measure of CNN for emotion detection} 
    \begin{tabular}{|l|c|}
\hline
\multicolumn{1}{|c|}{Metric} & Value \\ \hline
Accuracy                     & 95    \\ \hline
Precision                    & 89    \\ \hline
Recall                       & 79    \\ \hline
F1-Score                     & 98    \\ \hline
\end{tabular}
  \end{minipage}%
  \begin{minipage}{.6\columnwidth}
  \caption{Confusion metric of the proposed approach}
\label{Confusion metric of the proposed approach}
    \centering
   \begin{tabular}{|l|l|}
\hline
\multicolumn{1}{|c|}{Metric} & \multicolumn{1}{c|}{Value} \\ \hline
True Positive                & 15                         \\ \hline
False Positive               & 8                          \\ \hline
True negative                & 10                         \\ \hline
False Negative               & 7                          \\ \hline
\end{tabular}
  \end{minipage}
\end{table}
\section{Experimental result and analysis}
\label{Experimental result and analysis}
The module for acknowledgment consists of two phases:
\begin{itemize}
    \item the extraction of highlights to create a test informative index and
\item Combining:
CNN is used to describe an event of test data into an emotion class.
\end{itemize}
The CNN order is a fantastic grouping method. CNN's classification relies heavily on the premise that similar views belong in similar groupings.

By running our CNN model for about 50 epochs (considered to be a dataset passed forward and backward through CNN), we were able to gather information about the efficacy and accuracy of the model. Next, test images are used to evaluate the model. Additional assessment of the model is conducted for real-time emotion analysis using a range of input video and webcam sequences. The result of each frame is appropriately recorded, and any errors that occur during misclassification are also noted. As a result, the relevant measures are subsequently executed.
\begin{figure}
  \includegraphics[height= 5cm, width=\linewidth]{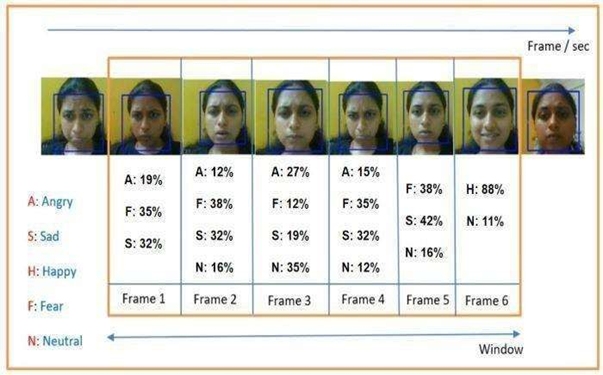}
  \caption{Basic emotion pattern recognition from a series of image.}
  \label{fig:Basic emotion pattern recognition from a series of image}
\end{figure}
\subsection{The CNN performance metric for identifying emotions}
The performance of the proposed system is shown in Table \ref{performance measure of CNN for emotion detection} and Table \ref{Confusion metric of the proposed approach}. From the Table \ref{performance measure of CNN for emotion detection}, Table \ref{Confusion metric of the proposed approach} and Fig \ref{fig:Basic emotion pattern recognition from a series of image}, it is concluded that the proposed model produces good performance for emotion detection.

\section{Conclusion}
The proposed study has demonstrated the effectiveness of ensemble models in forecasting dengue disease occurrences in the Chandigarh region of India. The primary objective of the work was to introduce a robust ensemble model for time series data forecasting, which can find applications in a wide range of disciplines beyond epidemiology. The work compared ensemble models with three established time series forecasting methods and found that they consistently outperformed the latter. The insights gained from this study can inform decision-making processes in public health, facilitate early intervention strategies, and contribute to more effective disease control and prevention efforts. In future, further refinement of ensemble models and the incorporation of additional data sources could lead to even more precise and timely disease forecasts.

\bibliographystyle{tfnlm}
\bibliography{interactnlmsample}

\end{document}